\documentclass[letterpaper,journal]{IEEEtran}

\usepackage{amsmath,amsfonts,amssymb,amsthm}
\usepackage[ruled,linesnumbered]{algorithm2e}
\usepackage{array}
\usepackage[caption=false,font=normalsize,labelfont=sf,textfont=sf]{subfig}
\usepackage{textcomp}
\usepackage{stfloats}
\usepackage{url}
\usepackage{verbatim}
\usepackage{graphicx}
\usepackage{cite}
\usepackage{booktabs}
\usepackage{multirow}
\usepackage{xcolor}
\usepackage{xspace}
\usepackage{pifont}
\usepackage{hyperref}
\hypersetup{
  colorlinks=true,linkcolor=blue,citecolor=blue,urlcolor=blue,
  pdftitle={TraceMark: Editor-Aware Watermarking for User Attribution in Text-Guided Image Editing},
  pdfauthor={Xiaodong Wu, Xiangman Li, Jianbing Ni, Rongxing Lu},
  pdfsubject={Image watermarking and user attribution in text-guided image editing},
  pdfkeywords={image watermarking, text-guided image editing, user attribution, multimedia provenance}
}
\graphicspath{{figures/}}
\newcommand{\tracemark}{\textsc{TraceMark}\xspace}

\newcommand{\R}{\mathbb{R}}

\DeclareMathOperator*{\argmax}{arg\,max}

\date{}

\begin{document}

\title{\tracemark: Editor-Aware Watermarking for User Attribution in Text-Guided Image Editing}

\author{Xiaodong~Wu,
        Xiangman~Li, \emph{Student Member, IEEE,}
        Jianbing~Ni, \emph{Senior Member, IEEE,} \\and Rongxing Lu, \emph{Fellow, IEEE,}%
        \thanks{X. Wu, X. Li, and J. Ni are with the Department of Electrical and Computer
        Engineering, Queen's University, Kingston, ON~K7L~3N6, Canada. E-mail:
        \{xiaodong.wu, xiangman.li, jianbing.ni\}@queensu.ca.}
        \thanks{R. Lu is with the School of Computing, Queen's University, Kingston, ON~K7L~3N6, Canada.
        E-mail: rongxing.lu@queensu.ca.}}

\maketitle
\thispagestyle{plain}

\begin{abstract}
Images shared outside an editing service can lose the information linking
them to the users who requested the edits. Recovering this link from the
image itself would support provenance tracking after redistribution, but a
user watermark must remain distinguishable after image content changes.
We propose \tracemark, an editor-aware watermarking framework that learns
to identify requesting users from edited images. A shared encoder embeds a
user-specific watermark before editing, and a shared detector identifies
the user by comparing image features with learned user reference vectors.
The encoder and
detector are trained jointly through an editor whose parameters remain
fixed. Errors in identifying the user guide the encoder toward watermarks
that remain recoverable after editing, while fidelity losses limit changes
to the input and intended output. Each instruction and image domain has a
separately trained model shared by its registered users. Controlled
experiments across three domains show that including the editor in training
improves user recovery with both reference-vector matching and binary
watermark decoding. The benefit therefore extends across both identification
methods tested.
Separately trained models achieve $87.5\%$ mean identification accuracy among
$500$ users. Tests that replace the editor without retraining the watermark
modules also recover user information, although accuracy depends on the
editor and input resolution. These findings support direct supervision of
user identification when learning watermarks for text-guided editing.
\end{abstract}


\section{Introduction}
\label{sec:intro}

\begin{figure}[t]
\centering
\includegraphics[width=0.95\linewidth]{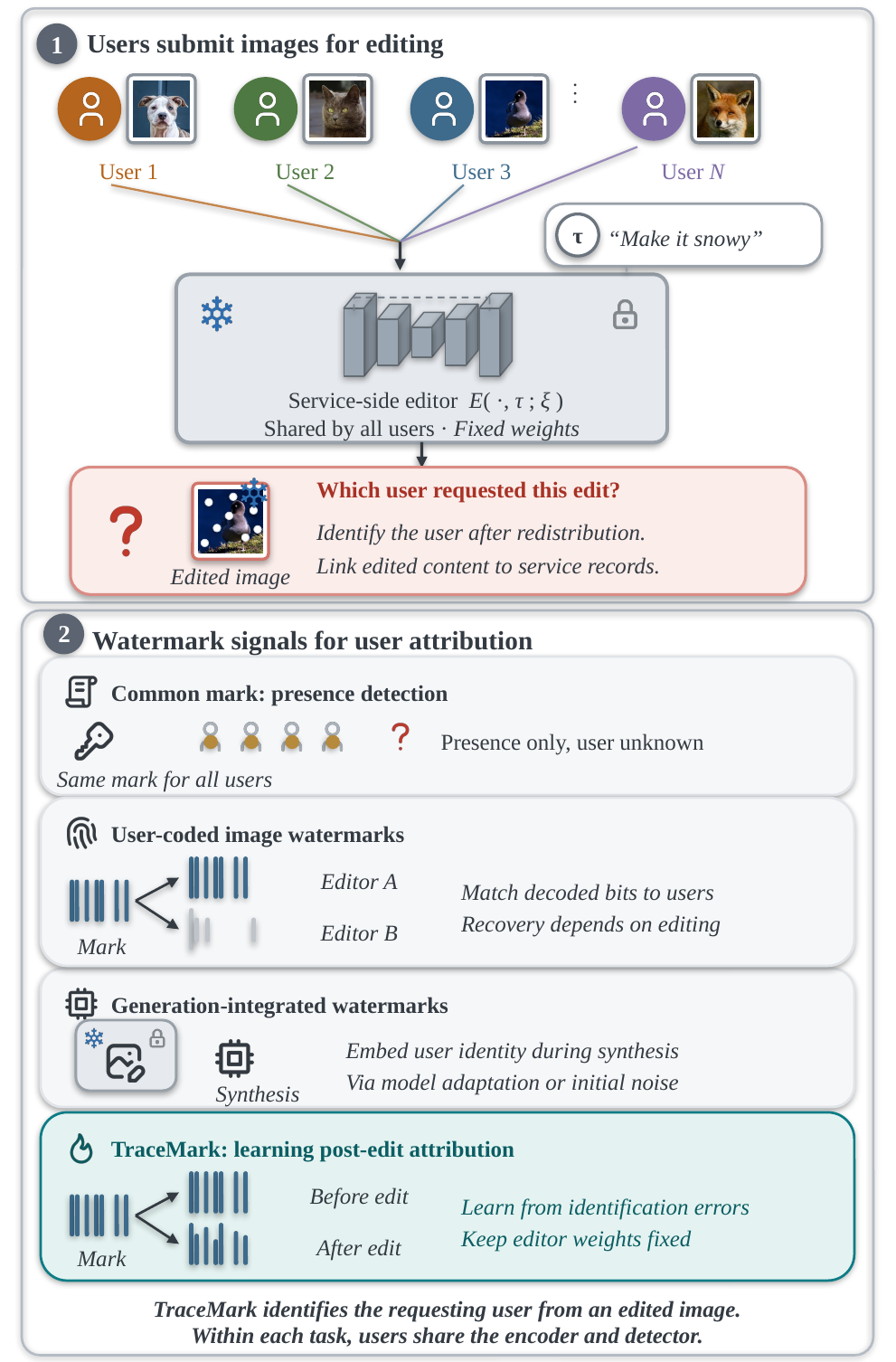}
\caption{System and watermark-integration paths for user attribution in
text-guided image editing.}
\label{fig:problem}
\end{figure}

\IEEEPARstart{T}{ext-guided} image editing allows users to change the
appearance and content of an existing image with a natural-language prompt.
Editors such as InstructPix2Pix~\cite{brooks2023instructpix2pix} can modify a
portrait in response to \emph{make the hair blond}, producing an image that
can subsequently be downloaded and shared. An editing service records who
requested the change, but this information may be lost when the result is
reposted outside the service. Recovering the requesting user from the image
would reconnect redistributed content with its provenance and allow the
service to consult the corresponding request records. This motivates the
user-attribution problem illustrated in Figure~\ref{fig:problem}: how can an
image editing service trace an edited image back to the user who requested
the edit?

A straightforward approach is to add a user-specific watermark after the
edit is complete. This can identify the user from the released image, but
does not by itself protect the association if that image is edited again.
The underlying challenge is to keep user information recoverable through
semantic changes~\cite{cao2026survey}. Watermarks trained to withstand
compression or resizing can be damaged when a generative editor reconstructs
image content, introducing errors into the recovered user information.
VINE~\cite{lu2024vine} and Robust-Wide~\cite{hu2024robustwide} improve
message recovery under editing, while SafeMark~\cite{wu2026safemark} trains
the editor to preserve an existing watermark. These methods support
attribution by assigning a different message to each user. Their training
targets message recovery, and the recovered message is then matched to the
registered users' messages. For attribution, the final goal is to distinguish
the requesting user from the other registered users. We therefore study
whether learning user-specific watermarks and this identification decision
jointly can improve recovery after editing. The evaluated image carries one
user's watermark and undergoes one edit.

We propose \tracemark, an editor-aware watermarking framework that learns
user-specific signals by directly optimizing user identification from edited
images. The service supplies the requesting user's identifier to a shared
encoder, which embeds the corresponding watermark in the input image.
After the editor applies the requested change, a shared detector converts
the edited image into a feature vector. The system identifies the user whose
learned reference vector is most similar to this vector. Training compares
this prediction with the requesting user's identifier and propagates the
resulting error through the fixed-weight editor to the encoder. This feedback
allows embedding and identification to be learned together from the edited
output. Fidelity losses additionally keep the marked input close to the
original image and its edit close to the result obtained without a watermark.
All users share the encoder and detector within a model trained for one
instruction and image domain. Once the appropriate task model is selected,
identification uses the edited image without requiring the original image.

Because attribution depends on what survives the editing process, the
evaluation first tests user identification with the editor used for training
and then replaces that editor while retaining the same watermark modules
(research question 1, RQ1). These tests assess recovery under both matched
and changed editing conditions. To determine how training contributes to
recovery, matched controls enable or disable training through the editor
for each of two identification methods: matching learned user reference
vectors and decoding user-specific binary messages (RQ2). We then vary
watermark distortion, edit magnitude, input resolution, editing instruction,
and user population
to examine the conditions under which attribution remains effective (RQ3).
The evaluated images come from the training image pools, so these results
do not establish recovery on previously unseen images.

Experiments across three image domains show that training through the
editor improves both identification approaches, while changing the
identification approach has a smaller and less consistent effect.
Separately trained models achieve $87.5\%$ mean accuracy among $500$ users.
Across $90$ editing tasks, \tracemark outperforms four released watermarks
on three unseen editing channels at a common 256-pixel resolution.
These gains have clear boundaries: RoSteALS is stronger on the editor used
to train \tracemark, VINE-R and Robust-Wide outperform the 256-pixel
\tracemark model at their released 512-pixel resolution, and \tracemark's
accuracy also drops substantially on ChordEdit.
The results support user-attribution training and establish the need to
validate recovery under the intended editing conditions.

\smallskip\noindent\textbf{Contributions.}
\begin{itemize}
    \item \textbf{User attribution through editing.}
          \tracemark jointly learns user-conditioned input watermarks and a
          shared detector to identify requesting users from edited images
          without updating the editor's parameters.

    \item \textbf{Evidence for the training mechanism.}
          \tracemark's matched controls establish consistent gains from
          training through the editor across two attribution approaches and
          three domains within a shared architecture.

    \item \textbf{Measured attribution capability.}
          \tracemark supports $500$-user identification and improves recovery
          on three unseen editing channels at a common carrier resolution,
          with fidelity and resolution controls that establish where these
          comparisons hold.
\end{itemize}

\smallskip\noindent
Sections~\ref{sec:related} and~\ref{sec:prelim} review prior work and develop
the motivating study. Section~\ref{sec:method} presents \tracemark,
Section~\ref{sec:exp} evaluates user attribution and its robustness, and
Section~\ref{sec:conclusion} concludes the paper.

\section{Related Work}
\label{sec:related}

\subsection{User Attribution with Image Watermarks}
\label{sec:related:image}
Assigning a different binary watermark to each user allows recovered messages
to identify users. Learned methods preserve these messages through simulated
distortions~\cite{zhu2018hidden}, physical print-and-capture
channels~\cite{tancik2020stegastamp,qin2024printcamera}, and video
transformations~\cite{zhang2019rivagan}. For existing images,
TrustMark~\cite{bui2023trustmark} supports variable resolutions, while
RoSteALS~\cite{bui2023rosteals} embeds messages through a pretrained
autoencoder. Both accept different messages with shared network weights,
allowing multi-user attribution without a separate encoder per user.
Section~\ref{sec:prelim} examines when this capability survives editing and
when bit errors make user identification ambiguous.

\label{sec:related:gen}
Generation-integrated methods instead associate images with users, keys, or
model recipients during synthesis. In latent diffusion
models~\cite{rombach2022ldm}, signatures can enter through decoder
fine-tuning~\cite{fernandez2023stablesignature}, user-conditioned
weights~\cite{kim2024wouaf}, or variable-message
conditioning~\cite{xiong2023flexible}. Latent
Watermark~\cite{meng2025latentwatermark} embeds and detects messages in
diffusion latent space. MarkPlugger~\cite{zhang2025markplugger} reuses
latent-space watermarks across generator variants without updating their
weights. Initial-noise methods provide another integration point for
detection or identification~\cite{wen2023treering,yang2024gaussianshading,ci2024ringid}.
These approaches establish attribution during generation. In image editing,
the additional question is whether the user signal remains recoverable when
an existing input is transformed by a requested semantic edit.

\subsection{Watermark Recovery Through Text-Guided Editing}
\label{sec:related:editing}
Text-guided editors reconstruct image content using noise injection,
attention control, or instruction conditioning~\cite{meng2022sdedit,hertz2023prompt2prompt,brooks2023instructpix2pix}.
Further advances in editing and its supervision~\cite{zhang2023magicbrush,geng2024instructdiffusion,sheynin2024emuedit}
broaden the transformations a watermark may encounter beyond conventional
compression and filtering.
A visually coherent edit can still replace image details that carry the
watermark. Recovery after compression or filtering therefore does not
establish that the same user signal will survive semantic editing.

VINE~\cite{lu2024vine} and Robust-Wide~\cite{hu2024robustwide} learn
watermarks for these editing conditions. VINE uses generative priors and
editing-related distortions, with VINE-R additionally trained on
InstructPix2Pix outputs. Robust-Wide includes instruction-driven editing in
watermark training. Both train watermark components with a fixed downstream
editor and report tests across editing configurations. Their objective is
message recovery, which supports attribution when messages are assigned to
users. \tracemark directly optimizes user identification, jointly learning
the embedded signal, detector, and user reference vectors used for matching.
Both the signal and the identification references are therefore learned from
post-edit user identification errors under the training editor.

SafeMark~\cite{wu2026safemark} approaches preservation from the editor side.
It fine-tunes the editor to retain an existing watermark, which may already
identify a source or user, while keeping the watermark encoder and decoder
fixed. \tracemark instead learns the requesting user's watermark and its
identification through a fixed editor. Thus, SafeMark adapts editing to an
existing signal, whereas \tracemark adapts the signal and user identification
to editing.

\subsection{Evaluating Attribution and Robustness}
\label{sec:related:robustness}
Robustness benchmarks~\cite{an2024waves} and studies of
regeneration~\cite{zhao2024invisible}, diffusion purification and model
substitution~\cite{saberi2024robustness}, and adaptive
attacks~\cite{lukas2024leveraging} show that recovery depends on the
transformation and attacker access. Withholding the decoder does not prevent
removal, as demonstrated by no-box attacks~\cite{wu2025nodecoder} and
MarkNull's model-agnostic latent manipulation~\cite{cao2026marknull}.
Consequently, attribution comparisons must specify the editing conditions
and measure correct user recovery alongside image fidelity. Our evaluation
covers requested edits in a trusted service and examines editor, resolution,
and population effects. Resistance to deliberate removal is outside this
evaluation of user attribution under requested edits.

\section{Motivating Study and Design Implications}
\label{sec:prelim}

This section studies whether released image watermarks can identify a
registered user after text-guided editing. We first assign different binary
messages to users and test whether the recovered messages identify them.
We then examine how bit errors affect identification under conventional
distortions, autoencoder reconstruction, and semantic editing. Finally,
we compare editors that change images by similar amounts to test whether
edit magnitude explains differences in attribution. The findings motivate the
training design in Section~\ref{sec:method} and its controlled evaluation in
Section~\ref{sec:exp:training-readout} across two attribution methods within
the same watermark architecture.

\subsection{Existing Watermarks Support Attribution Under Compatible Editors}
\label{sec:motivation:feasibility}

A $k$-bit watermark can identify a user from a fixed registry by assigning
each user $u$ a distinct binary message, or codeword,
$\mathbf{c}_u\in\{0,1\}^{k}$. After a marked image is transformed, the
decoder returns a recovered message $\hat{\mathbf{c}}$. We select the user
whose assigned message differs from it in the fewest bit positions, measured
by Hamming distance. For all $100$-bit methods, we use the same codebook,
whose messages are chosen to differ in many bit positions. The codebook is
fixed across images, editing features, and datasets. If multiple users are
equally close to the decoded message, the system has no unique prediction and
we record the result as an attribution error. We evaluate all $90$ editing
features, each corresponding to a target editing instruction, with $100$
images per feature and $100$ registered users.

Figure~\ref{fig:channel-ladder}(a) shows that the three released watermarks
identify users almost perfectly before transformation (the identity channel)
and after compression or blur (the pixel-domain channels).
RoSteALS also reaches $0.975$ after InstructPix2Pix (IP2P), establishing that
an image watermark can support user attribution after a compatible
editor. The same panel shows much lower accuracy under other editing channels.
A well-separated registry therefore enables attribution but cannot compensate
when a channel changes too much of the embedded message. We next quantify this
decoding limit for a fixed population of registered users.

\subsection{Capacity Margin Links Bit Recovery to Attribution}
\label{sec:motivation:capacity}

The preceding results raise a practical question: how much bit corruption can
a registry tolerate before user identities become ambiguous? Bit accuracy
alone is incomplete because the same error rate can support different registry
sizes when message lengths differ. Distinguishing $N$ users requires
$\log_2N$ bits, which is $6.64$ bits for the $100$-user registry used here.
For a $k$-bit watermark with observed bit error rate $p$, we approximate
decoding errors with a binary symmetric channel, in which each bit flips
independently with probability $p$. The resulting information capacity is
$k[1-H_2(p)]$, where
$H_2(p)=-p\log_2p-(1-p)\log_2(1-p)$ is binary entropy~\cite{shannon1948mathematical}.
We measure the surplus over the registry requirement with
\begin{equation}
\label{eq:capacity-margin}
M(k,p,N)=k\bigl[1-H_2(p)\bigr]-\log_2N.
\end{equation}
A positive $M$ means that the estimated remaining information exceeds what the
registry needs. A negative $M$ indicates an information deficit. Because
editing can introduce correlated bit errors, $M$ is a diagnostic quantity and
does not guarantee a particular identification accuracy.

\begin{figure*}[t]
\centering
\includegraphics[width=0.95\textwidth]{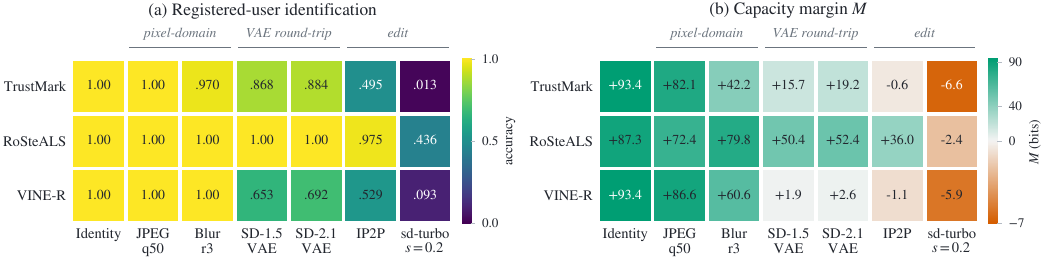}
\caption{Registered-user identification (a) and capacity margin from bit
recovery (b) for three released $100$-bit watermarks.}
\label{fig:channel-ladder}
\end{figure*}

\begin{table}[t]
\centering
\caption{Registered-user identification under InstructPix2Pix and Stable
Diffusion XL (SDXL)-Turbo.}
\label{tab:motivation-feasibility}
\footnotesize
\setlength{\tabcolsep}{3.5pt}
\begin{tabular*}{\columnwidth}{@{\extracolsep{\fill}}lccc@{}}
\toprule
Method & \#Bits & IP2P & SDXL-Turbo \\
\midrule
TrustMark~\cite{bui2023trustmark}   & 100 & 0.495 & 0.009 \\
RoSteALS~\cite{bui2023rosteals}     & 100 & \textbf{0.975} & 0.249 \\
VINE-R~\cite{lu2024vine}            & 100 & 0.529 & \textbf{0.922} \\
Robust-Wide~\cite{hu2024robustwide} & 64  & 0.917 & 0.648 \\
\bottomrule
\end{tabular*}
\end{table}

Figure~\ref{fig:channel-ladder}(b) reports this margin for each method and
transformation in panel (a). A channel here is the transformation between
watermark embedding and decoding. The three channel groups cover conventional
pixel-domain processing, autoencoder round trips that encode and reconstruct
an image, and semantic editing. Identity, Joint Photographic Experts Group (JPEG)
compression, and Gaussian blur leave at least
$42.2$ bits of margin for every method, together with identification accuracy
of at least $0.970$. The autoencoder round trips retain ample margin for TrustMark
and RoSteALS but place VINE-R close to the registry requirement. IP2P leaves
RoSteALS with a $36.0$-bit margin while TrustMark and VINE-R are near zero.
Under sd-turbo, all three margins become negative and identification accuracy
falls to $0.013$ for TrustMark, $0.436$ for RoSteALS, and $0.093$ for VINE-R.

The progression from large positive margins to information deficits illustrates
why successful message recovery from an unedited marked image does not
establish post-edit attribution.
It motivates training on the user identity recovered from the edited output,
so the watermark receives feedback about the information that survives the
target transformation. Section~\ref{sec:exp:training-readout} tests the effect
of this training signal with a matched control.

\subsection{Editor Compatibility Shapes Attribution}
\label{sec:motivation:backbone}

The preceding experiment links bit errors to identification errors, but it
does not explain why methods respond differently to two semantic editors. A
method may fail because an editor changes the image more, or because the
editor is incompatible with the model used to construct the watermark. We
test whether edit magnitude alone explains the difference by measuring each
editor on unmarked covers with a
fixed Stable Diffusion 1.5 variational autoencoder (VAE). The normalized latent
displacement is
$\lVert\mathbf{z}_{\mathrm{edit}}-\mathbf{z}_{\mathrm{cover}}\rVert_2/
\lVert\mathbf{z}_{\mathrm{cover}}\rVert_2$, where
$\mathbf{z}_{\mathrm{cover}}$ and $\mathbf{z}_{\mathrm{edit}}$ are the VAE
representations of the input and edited image. Using the same VAE provides a
common measure of image change across editors.

We compare IP2P with Stable Diffusion XL (SDXL)-Turbo at strength $0.2$.
The two editors produce nearly equal displacement, $0.683$ and $0.698$, over
all $90$ features and $50$
covers per feature. Despite this similar edit magnitude, the leading method
changes in Table~\ref{tab:motivation-feasibility}. RoSteALS embeds in the latent
space of a latent-diffusion autoencoder~\cite{rombach2022ldm} from the same
model family as the Stable Diffusion 1.5 base of IP2P. It falls from $0.975$
on IP2P to $0.249$ on SDXL-Turbo. VINE-R moves in the opposite direction, from
$0.529$ to $0.922$, and its encoder is built on SDXL-Turbo. Robust-Wide is
trained through IP2P and is also stronger on that editor. TrustMark has no
generative base model and is weak on both editors. Similar latent displacement
therefore does not imply similar attribution accuracy. The relationship
between performance and model family suggests that editor compatibility
also matters, although this comparison does not isolate its cause.

These observations motivate studying whether joint watermark encoder and detector
training through a target editor improves user attribution. \tracemark
implements this approach while keeping editor parameters fixed. The
evaluation separately measures performance across editors and the effect of
including the editor in training within the same architecture. A matched
control in Section~\ref{sec:exp:training-readout} measures this training effect
under prototype matching and binary decoding on one feature from each image
domain with identical training conditions.

\section{The \tracemark Framework}
\label{sec:method}

\begin{figure*}[t]
\centering
\includegraphics[width=0.95\linewidth]{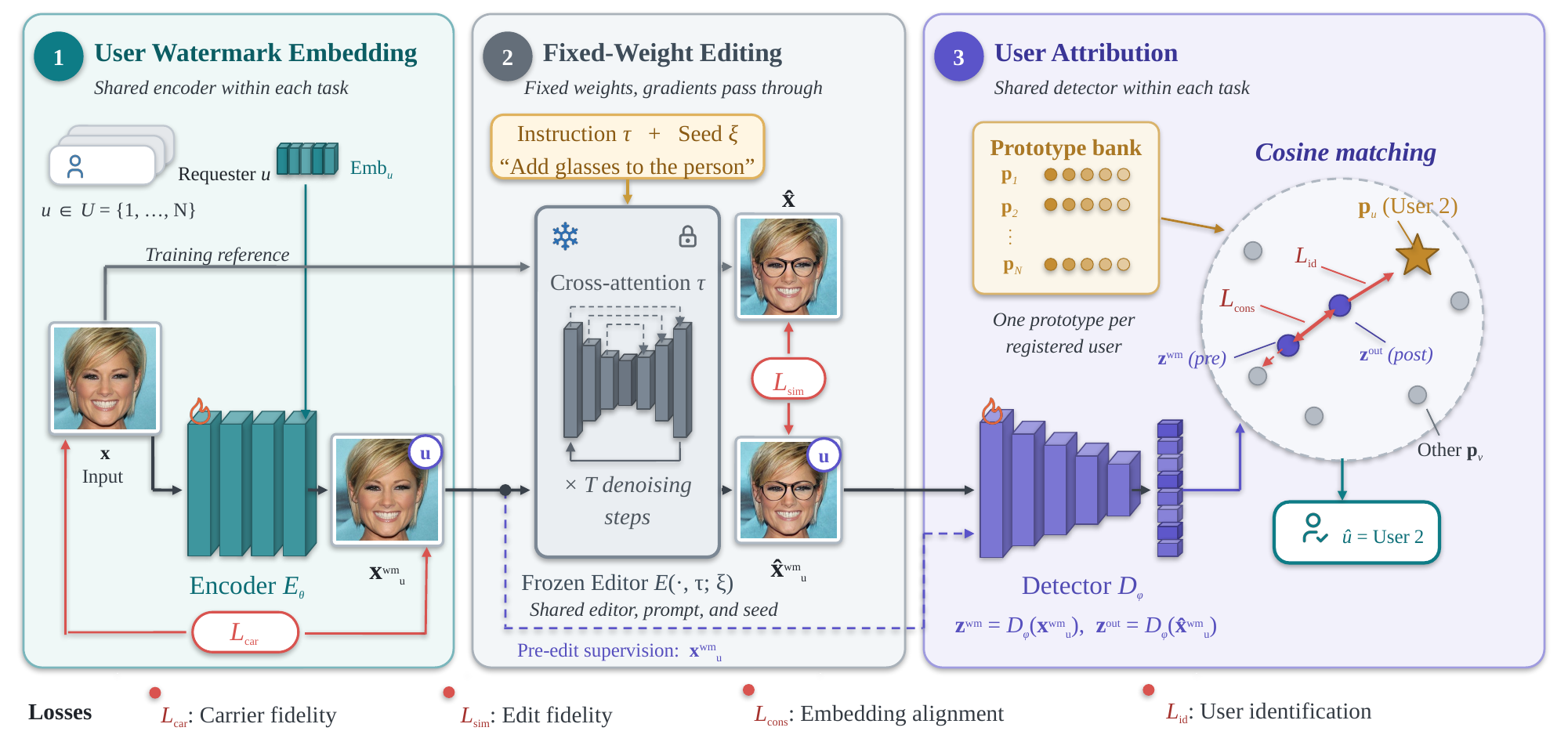}
\caption{The \tracemark training loop for user-conditioned watermark
embedding and attribution through a fixed-weight instruction editor.}
\label{fig:pipeline}
\end{figure*}

\subsection{Operational Setting and Design}
\label{sec:method:overview}
\tracemark learns to identify the requesting user from an edited image.
The service binds each request to its authenticated user and passes that
user's identifier, together with the input image, to the watermark encoder.
Its user-conditioned encoder $E_\theta$ produces a watermarked input
image, also called the marked carrier. Its detector $D_\phi$ maps an image
$\mathbf{y}$ to a feature vector, or embedding,
$D_\phi(\mathbf{y})\in\mathbb{R}^{d}$, where $d$ is the embedding dimension
and $\theta$ and $\phi$ are trainable parameters. The detector compares this
embedding with $N$ learned reference vectors, one per registered user. These
reference vectors are the user prototypes. Training jointly optimizes the
encoder, detector, and prototypes while keeping editor parameters fixed.
Prototype matching removes
the need to construct a binary registry codebook for the registered user population.

Let $\mathcal{U}=\{1,\ldots,N\}$ denote the registered user population. The
index $u\in\mathcal{U}$ identifies the user who requested the edit, while
$v\in\mathcal{U}$ indexes a candidate user with prototype
$\mathbf{p}_v\in\mathbb{R}^{d}$. Given a normalized red, green, and blue (RGB) cover image
$\mathbf{x}\in[-1,1]^{3\times H\times W}$ of height $H$ and width $W$, the
encoder produces
$\mathbf{x}^{\mathrm{wm}}_u=E_\theta(\mathbf{x},u)$. A fixed-weight editor
$\mathcal{E}$ takes an instruction $\tau$ and sampling seed $\xi$, producing
the unmarked and marked edits
$\hat{\mathbf{x}}=\mathcal{E}(\mathbf{x},\tau;\xi)$ and
$\hat{\mathbf{x}}^{\mathrm{wm}}_u=
\mathcal{E}(\mathbf{x}^{\mathrm{wm}}_u,\tau;\xi)$. A hat over an image thus
denotes the result after editing. For nonzero vectors $\mathbf{a}$ and
$\mathbf{b}$, let
$\cos(\mathbf{a},\mathbf{b})=\mathbf{a}^{\top}\mathbf{b}/
(\lVert\mathbf{a}\rVert_2\lVert\mathbf{b}\rVert_2)$. The predicted user is
\begin{equation}
\label{eq:goal}
\hat{u}=\argmax_{v\in\mathcal{U}}
\cos\!\left(D_\phi(\hat{\mathbf{x}}^{\mathrm{wm}}_u),
\mathbf{p}_v\right).
\end{equation}
Here, $\hat{u}$ denotes the predicted user. The true user $u$ supplies the
training label but is unknown to the detector at inference. Training seeks
$\hat{u}=u$. The marked carrier must remain visually close to $\mathbf{x}$,
and its edit $\hat{\mathbf{x}}^{\mathrm{wm}}_u$ must remain close to the
unmarked edit $\hat{\mathbf{x}}$ under the same instruction and seed.

Each model is trained for one domain,
instruction, editor, registry, and carrier resolution. Attribution uses
the corresponding detector and prototype bank, selected in advance.
Automatic task-model selection is outside the evaluation.

A trusted editing service controls the encoder, detector, prototype bank,
and user assignment. Equation~\ref{eq:goal} always selects a registered user,
even for an unmarked image. Deployment beyond this registered-user evaluation
therefore requires a calibrated rejection rule for unmarked images and
unknown users.
The recovered identity refers to the requesting user, whose relationship to
the image is established by the service's request record. It does not imply
image ownership or authorship. Since the marked input already carries this
identity, extraction alone also cannot certify that editing occurred.
Adaptive removal, watermark forgery, and leakage of watermark components
remain outside the evaluation.

Embedding before the requested edit lets us evaluate whether the user signal
survives that transformation. A service that controls the final output can
also embed after editing, although that placement alone gives no assurance
of recovery after a subsequent edit. Our experiments measure recovery through
the editor and do not compare the two placements. Attribution refers to the
user associated with the evaluated request and does not reconstruct the
sequence of users in a multi-stage editing history.

\subsection{Architecture}
\label{sec:method:arch}
The encoder and detector weights are shared across all $N$ users, while each
user has two learned vectors with different roles: an embedding that tells
the encoder which user's watermark to write, and a prototype used for
identification. Applying the same detector before and after editing makes
the resulting embeddings directly comparable. The consistency objective
uses this comparison to penalize edit-induced movement of the image
representation within the shared detector embedding space used for user
identification.

\paragraph{Identifier encoder}
Let $\{\mathrm{Emb}_u\in\R^{d_u}\}_{u\in\mathcal U}$ be the learned user
embedding table within $E_\theta$, where $d_u$ is the user-embedding
dimension. Given an input image
$\mathbf{x}\in[-1,1]^{3\times H\times W}$, the encoder broadcasts
$\mathrm{Emb}_u$ over the $H\times W$ spatial grid and concatenates it with
the image along the channel axis. A small fully convolutional stack outputs
a residual field
$\delta_\theta(\mathbf{x},\mathrm{Emb}_u)\in\R^{3\times H\times W}$.
The fixed carrier strength $\alpha>0$ scales this field after a $\tanh$
activation:
\begin{equation}
\label{eq:encoder}
\mathbf{x}^{\mathrm{wm}}_u
\;=\;\mathrm{clamp}\,\Bigl(\mathbf{x}+\alpha\,\tanh\,\bigl(
\delta_\theta(\mathbf{x},\,\mathrm{Emb}_u)\bigr),\,-1,\,1\Bigr).
\end{equation}
The clamp acts elementwise. Since the output of $\tanh$ lies in $[-1,1]$,
$\alpha$ bounds each absolute pixel residual before clamping and controls the
tradeoff between carrier fidelity and identification capacity evaluated in
Section~\ref{sec:exp:fidelity-control}.

\paragraph{Fixed-weight editor}
During training, the editor remains in evaluation mode with parameter
gradients disabled. Its output-to-input computation graph is retained.
This access lets gradients reach $E_\theta$ through the marked edit
without updating the editor parameters. An editor accessible only through
a remote service cannot be used for this training path. We use a fixed sampler and
inference-step budget for reproducible editor execution, and run the editor
subgraph in single precision for numerical stability. At deployment, the
trained modules operate without gradient computation.

\paragraph{Detector and attribution}
The detector $D_\phi$ is a compact strided convolutional backbone
followed by global average pooling and a linear head that produces
an embedding $\mathbf{z}\in\R^{d}$. For attribution, its input is the edited
image, so $\mathbf{z}=D_\phi(\hat{\mathbf{x}}^{\mathrm{wm}}_u)$.
For any nonzero vector $\mathbf{a}$,
let $\bar{\mathbf{a}}=\mathbf{a}/\lVert\mathbf{a}\rVert_2$ denote its
$\ell_2$-normalized form. The classification scores, or logits, are scaled
cosine similarities:
\begin{equation}
\label{eq:logits}
\boldsymbol{\ell}(\mathbf{z})
\;=\;s\;\bigl[\bar{\mathbf{p}}_1^{\top}\bar{\mathbf{z}},\,\ldots,\,
\bar{\mathbf{p}}_N^{\top}\bar{\mathbf{z}}\bigr]
\;\in\;\R^N,
\end{equation}
where $s>0$ is a learned logit scale capped at $100$, and
$\boldsymbol{\ell}(\mathbf{z})_v$ is the classification logit for candidate
user $v$. Attribution selects the prototype with the highest cosine similarity,
$\hat{u}=\argmax_{v\in\mathcal{U}}\,\boldsymbol{\ell}(\mathbf{z})_v$.
Exact channel widths, the value of $\alpha$, the detector dimensions,
and the logit-scale configuration are reported in Appendix~\ref{app:impl}.

\subsection{Training Objective}
\label{sec:method:loss}
For the paired images defined in Section~\ref{sec:method:overview}, the detector produces
$\mathbf{z}^{\mathrm{wm}}=D_\phi(\mathbf{x}^{\mathrm{wm}}_u)$ and
$\mathbf{z}^{\mathrm{out}}=D_\phi(\hat{\mathbf{x}}^{\mathrm{wm}}_u)$ before
and after editing. Image-space $\ell_1$ distances are measured after mapping
images to $[0,1]$ and are averaged over pixels and channels.
At training round $t$, the edit-loss ramp
$r_t=\mathrm{clip}((t-t_{\mathrm{warm}})/t_{\mathrm{ramp}},0,1)$ is set by
warm-up length $t_{\mathrm{warm}}$ and ramp length $t_{\mathrm{ramp}}$, both
measured in training rounds. It increases the weights of edit-dependent
losses from zero to their full values.
Four losses target carrier and edit fidelity, embedding consistency, and
user attribution.

\emph{Carrier fidelity.} We penalize the residual added to the input image:
\begin{equation}
\label{eq:car}
\mathcal{L}_{\mathrm{car}}
= \|\mathbf{x}^{\mathrm{wm}}_u - \mathbf{x}\|_1 .
\end{equation}
The hard bound in Eq.~\ref{eq:encoder} limits the largest pixel change, while
$\mathcal{L}_{\mathrm{car}}$ reduces the average residual within that bound.

\emph{Edit fidelity.} We keep the marked edit close to the corresponding
unmarked edit generated with the same instruction and seed:
\begin{equation}
\label{eq:sim}
\mathcal{L}_{\mathrm{sim}}
= \|\hat{\mathbf{x}}^{\mathrm{wm}}_u - \hat{\mathbf{x}}\|_1 .
\end{equation}
This term constrains changes attributable to the watermark after the requested
transformation has been applied.

\emph{Embedding consistency.} We align the detector embeddings before and
after editing:
\begin{equation}
\label{eq:cons}
\mathcal{L}_{\mathrm{cons}}
= 1-\cos(\mathbf{z}^{\mathrm{wm}},\,\mathbf{z}^{\mathrm{out}}) .
\end{equation}
It encourages the two embeddings to point in the same direction, preserving
the representation used for cosine-based attribution across the registered
user population.

\emph{User attribution.} We apply cross-entropy supervision before and after
editing:
\begin{equation}
\label{eq:id}
\begin{aligned}
\mathcal{L}_{\mathrm{pre}}
&= \mathrm{CE}\!\left(\boldsymbol{\ell}(\mathbf{z}^{\mathrm{wm}}),u\right),\\
\mathcal{L}_{\mathrm{out}}
&= \mathrm{CE}\!\left(\boldsymbol{\ell}(\mathbf{z}^{\mathrm{out}}),u\right),\\
\mathcal{L}_{\mathrm{id}}
&= \mathcal{L}_{\mathrm{pre}}+\lambda_{\mathrm{out}}r_t\mathcal{L}_{\mathrm{out}}.
\end{aligned}
\end{equation}
Here, $\mathrm{CE}$ denotes cross entropy, and $\lambda_{\mathrm{out}}>0$
weights post-edit attribution relative to its pre-edit counterpart. The
$\mathcal{L}_{\mathrm{out}}$ term is the direct training signal for the final
decision in Eq.~\ref{eq:goal}.

The full objective combines the four losses:
\begin{equation}
\label{eq:loss}
\mathcal{L}
\;=\;\lambda_{\mathrm{id}}\mathcal{L}_{\mathrm{id}}
+\lambda_{\mathrm{sim}}\,r_t\,\mathcal{L}_{\mathrm{sim}}
+\lambda_{\mathrm{car}}\mathcal{L}_{\mathrm{car}}
+\lambda_{\mathrm{cons}}\,r_t\,\mathcal{L}_{\mathrm{cons}} .
\end{equation}
The nonnegative $\lambda$ coefficients balance the four losses. Carrier
fidelity and pre-edit attribution remain active throughout training. The
ramp progressively activates edit fidelity, embedding consistency, and
post-edit attribution, the last through Eq.~\ref{eq:id}.

\begin{algorithm}[t]
\caption{\tracemark training, with the minibatch loop omitted.}
\label{alg:train}
\KwIn{covers $\mathcal{X}$, users $\mathcal{U}$ ($|\mathcal{U}|{=}N$), target instruction $\tau$, training rounds $T$, warm-up $t_{\mathrm{warm}}$, ramp length $t_{\mathrm{ramp}}$}
\KwData{encoder $E_\theta$, detector $D_\phi$, prototypes $\{\mathbf{p}_u\}$, fixed-weight editor $\mathcal{E}$}
\KwOut{trained $E_\theta,D_\phi,\{\mathbf{p}_u\}$}
Initialize $\theta,\phi,\{\mathbf{p}_u\}$\;
\For{$t \gets 1$ \KwTo $T$}{
  Select cover $\mathbf{x}$, user $u$, and seed $\xi$\;
  $r_t \gets \mathrm{clip}\bigl((t-t_{\mathrm{warm}})/t_{\mathrm{ramp}},\,0,\,1\bigr)$ \tcp*{ramp}
  $\mathbf{x}^{\mathrm{wm}}_u \gets E_\theta(\mathbf{x},u)$ \tcp*{encode, Eq.~\ref{eq:encoder}}
  $\mathbf{z}^{\mathrm{wm}} \gets D_\phi(\mathbf{x}^{\mathrm{wm}}_u)$\;
  $\mathcal{L}_{\mathrm{car}} \gets \|\mathbf{x}^{\mathrm{wm}}_u-\mathbf{x}\|_1$ \tcp*{carrier fidelity}
  $\mathcal{L}_{\mathrm{id}} \gets \mathrm{CE}(\boldsymbol{\ell}(\mathbf{z}^{\mathrm{wm}}),u)$ \tcp*{pre-edit}
  \eIf{$r_t>0$}{
    $\hat{\mathbf{x}} \gets \mathcal{E}(\mathbf{x},\tau;\xi)$ \tcp*{edit clean}
    $\hat{\mathbf{x}}^{\mathrm{wm}}_u \gets \mathcal{E}(\mathbf{x}^{\mathrm{wm}}_u,\tau;\xi)$ \tcp*{edit marked}
    $\mathbf{z}^{\mathrm{out}} \gets D_\phi(\hat{\mathbf{x}}^{\mathrm{wm}}_u)$\;
    $\mathcal{L}_{\mathrm{sim}} \gets \|\hat{\mathbf{x}}^{\mathrm{wm}}_u-\hat{\mathbf{x}}\|_1$\;
    $\mathcal{L}_{\mathrm{cons}} \gets 1-\cos(\mathbf{z}^{\mathrm{wm}},\mathbf{z}^{\mathrm{out}})$\;
    $\mathcal{L}_{\mathrm{id}} \mathrel{+}= \lambda_{\mathrm{out}} r_t\,\mathrm{CE}(\boldsymbol{\ell}(\mathbf{z}^{\mathrm{out}}),u)$ \tcp*{post-edit}
  }{
    $\mathcal{L}_{\mathrm{sim}},\mathcal{L}_{\mathrm{cons}} \gets 0$ \tcp*{warm-up: skip $\mathcal{E}$}
  }
  $\mathcal{L} \gets \lambda_{\mathrm{id}}\mathcal{L}_{\mathrm{id}}+\lambda_{\mathrm{sim}}r_t\mathcal{L}_{\mathrm{sim}}+\lambda_{\mathrm{car}}\mathcal{L}_{\mathrm{car}}+\lambda_{\mathrm{cons}}r_t\mathcal{L}_{\mathrm{cons}}$\;
  Update $\theta,\phi,\{\mathbf{p}_u\}$ by a gradient step on $\mathcal{L}$ \tcp*{$\mathcal{E}$ frozen}
}
\end{algorithm}

\subsection{Training Procedure}
\label{sec:method:algo}
Algorithm~\ref{alg:train} outlines the update for one image, omitting the
inner minibatch loop. In the implementation, each training round processes
a fixed image quota in minibatches. The ramp $r_t$ advances once per round
and is shared by its minibatches (lines~3--4). During warm-up, carrier
fidelity and pre-edit attribution train the encoder to embed user information
and the detector to recover it from unedited marked images (lines~5--8).
While $r_t=0$, the editor is skipped and the edit-dependent terms remain zero
(line~17).

Once $r_t>0$, the same instruction and seed edit the unmarked and marked
inputs (lines~10--11). Their outputs define edit fidelity, embedding
consistency, and post-edit attribution (lines~13--15). Increasing $r_t$
increases the influence of these objectives as training proceeds. The combined loss in
Eq.~\ref{eq:loss} updates the encoder, detector, and prototypes (lines~19--20).

\subsection{Complexity and Margin Analysis}
\label{sec:method:analysis}
Let $\bar{\mathbf z}$ and $\bar{\mathbf z}'$ be the $\ell_2$-normalized detector
embeddings of a marked image before and after editing or degradation.
For candidate user $v$, let
$c_v(\bar{\mathbf z})=\bar{\mathbf p}_v^{\top}\bar{\mathbf z}$ denote the
unscaled cosine score. For an image assigned to user $u$, define the
attribution margin as the correct user's score minus the highest competing
user score:
\begin{equation}
\label{eq:margin}
\gamma_u(\bar{\mathbf z})=c_u(\bar{\mathbf z})-
\max_{v\ne u}c_v(\bar{\mathbf z}).
\end{equation}
Suppose the transformation satisfies
$\lVert\bar{\mathbf z}'-\bar{\mathbf z}\rVert_2\le\epsilon$. Since
$\lVert\bar{\mathbf p}_v\rVert_2=1$, the Cauchy--Schwarz inequality gives
$|c_v(\bar{\mathbf z}')-c_v(\bar{\mathbf z})|\le\epsilon$ for every candidate.
The correct-user score can therefore decrease by at most $\epsilon$, while
the largest competing score can increase by at most $\epsilon$. The predicted
user remains $u$ whenever
$\gamma_u(\bar{\mathbf z})>2\epsilon$.

\begin{table*}[t]
\centering
\caption{Cross-editor identification in three image domains using a common
256-pixel pipeline. Each domain contains 30 editing features.}
\label{tab:cross-editor-resolution}
\small
\renewcommand{\arraystretch}{1.02}
\setlength{\tabcolsep}{5pt}
\begin{tabular*}{0.98\textwidth}{@{\extracolsep{\fill}}llccccc@{}}
\toprule
Editing channel & Domain & \tracemark & TrustMark & RoSteALS & VINE-R & Robust-Wide \\
\midrule
\multirow{3}{*}{InstructPix2Pix, training channel}
 & CelebA-HQ & 0.901 & 0.554 & \textbf{0.984} & 0.677 & 0.932 \\
 & AFHQ-Dog  & 0.806 & 0.320 & \textbf{0.955} & 0.376 & 0.884 \\
 & CUB-Bird  & 0.891 & 0.611 & \textbf{0.986} & 0.533 & 0.934 \\
\midrule
\multirow{3}{*}{sd-turbo, $s=0.2$}
 & CelebA-HQ & \textbf{0.877} & 0.019 & 0.355 & 0.140 & 0.255 \\
 & AFHQ-Dog  & \textbf{0.864} & 0.012 & 0.422 & 0.069 & 0.291 \\
 & CUB-Bird  & \textbf{0.775} & 0.009 & 0.531 & 0.071 & 0.427 \\
\midrule
\multirow{3}{*}{sd-turbo, p2p-style $s=0.3$}
 & CelebA-HQ & \textbf{0.790} & 0.009 & 0.102 & 0.072 & 0.110 \\
 & AFHQ-Dog  & \textbf{0.783} & 0.008 & 0.138 & 0.036 & 0.142 \\
 & CUB-Bird  & \textbf{0.656} & 0.010 & 0.162 & 0.030 & 0.284 \\
\midrule
\multirow{3}{*}{ChordEdit}
 & CelebA-HQ & \textbf{0.427} & 0.009 & 0.061 & 0.038 & 0.034 \\
 & AFHQ-Dog  & \textbf{0.271} & 0.007 & 0.067 & 0.017 & 0.039 \\
 & CUB-Bird  & \textbf{0.230} & 0.007 & 0.057 & 0.019 & 0.070 \\
\bottomrule
\end{tabular*}
\end{table*}

This bound is a sufficient condition for preserving one image's predicted
user. Applying it to a given
editor or degradation requires measuring or bounding its embedding
displacement $\epsilon$. The identification loss promotes a larger margin,
and the consistency loss penalizes displacement caused by the target edit.
The bound explains why both score separation and embedding stability matter
for attribution. It does not guarantee recovery under an arbitrary editor.
Attributing improved recovery to a design choice requires
a matched control. Section~\ref{sec:exp:training-readout} isolates the training
path and attribution method under matched conditions.

For one trained population, the prototype bank and user-embedding table
require $O\bigl(N(d+d_u)\bigr)$ storage. After the detector produces one embedding, forming all
user logits costs $O(Nd)$ during training or inference. The other network
parameters and computations are independent of $N$. The current implementation retrains its
learnable components when the registry expands.
Population-specific results appear in Section~\ref{sec:exp:population-capacity}
for registries of up to $500$ users.

\section{Experiments}
\label{sec:exp}

\subsection{Experimental Setup}
\label{sec:exp:setup}

\paragraph{Datasets and editing features}
Each editing feature specifies a target instruction, such as changing hair
color or adding glasses. We evaluate $30$ features in each of three multimedia
domains: high-quality CelebA (CelebA-HQ) faces~\cite{karras2018progressive},
the Animal Faces-HQ (AFHQ) Dog split~\cite{choi2020stargan}, and
Caltech-UCSD Birds-200-2011 (CUB-200-2011)~\cite{wah2011cub}. Images are
cropped to $256\times256$ as detailed in Appendix~\ref{app:data}.
Each \tracemark model is trained for one of the
$90$ domain-feature settings through InstructPix2Pix~\cite{brooks2023instructpix2pix}.
Cross-editor tests replace the editing channel while keeping the encoder,
detector, and prototype bank fixed.

\paragraph{Compared watermarks}
We compare \tracemark with the released TrustMark~\cite{bui2023trustmark},
RoSteALS~\cite{bui2023rosteals}, VINE-R~\cite{lu2024vine}, and
Robust-Wide~\cite{hu2024robustwide} checkpoints (Appendix~\ref{app:baselines}).
For $N=100$, the first three
baselines receive distinct $100$-bit codewords from a balanced codebook, while
Robust-Wide uses its released $64$-bit messages. Nearest-codeword decoding
maps each output to a user. If several codewords are equally close, the
prediction is ambiguous and is counted as an error. The main comparison
uses 256-pixel carriers, the native resolution of \tracemark, TrustMark, and
RoSteALS. The control in Appendix~\ref{app:resolution-domain} retains the released 512-pixel
resolution of VINE-R and Robust-Wide. Changing \tracemark's resolution
requires retraining.

\paragraph{Editing channels and evaluation scope}
The four channels in RQ1 are the InstructPix2Pix training channel, sd-turbo
image-to-image editing~\cite{sauer2023sdturbo} at strength $0.2$, sd-turbo
editing at strength $0.3$ with a pix2pix-style configuration, and
ChordEdit~\cite{chordedit2026}. Table~\ref{tab:cross-editor-resolution} reports
the mean identification accuracy over the $30$ features in each domain.
RQ2 uses one feature from each
domain. RQ3 evaluates all settings for the fidelity control, cross-instruction
reuse in all three domains, and four CelebA-HQ tasks for registries from
$100$ to $500$ users. Appendix~\ref{app:impl}
gives detailed sample counts and hyperparameters.
Paired comparisons reuse the user population, covers, instructions, and editor
seeds. The evaluated \tracemark carriers come from the final training
round, and the baselines use the same covers. These comparisons use images
from the training pools. They do not measure generalization to held-out images.

\paragraph{Metrics}
Identification accuracy is the fraction of marked outputs assigned to the
correct registered user. For bitstring methods, bit accuracy is the fraction
of correctly decoded watermark bits. Carrier fidelity uses peak signal-to-noise
ratio (PSNR), structural similarity index measure (SSIM)~\cite{wang2004ssim},
and learned perceptual image patch similarity (LPIPS)~\cite{zhang2018lpips}.
Section~\ref{sec:motivation:backbone} also measures normalized latent
displacement with a fixed Stable Diffusion 1.5 variational autoencoder for
both evaluated editors.

\subsection{Post-Edit Attribution (RQ1): Matched and Unseen Editors}
\label{sec:exp:gap}
\label{sec:exp:cross-editor}

\textbf{At 256 pixels, \tracemark leads on the three unseen editing channels.}
Table~\ref{tab:cross-editor-resolution} shows this advantage in every image
domain. Averaged across domains, \tracemark reaches $1.9$, $4.2$, and $5.0$
times the strongest released baseline on sd-turbo, its p2p-style
configuration, and ChordEdit, respectively. Released watermarks nevertheless
perform well on the InstructPix2Pix training channel, where RoSteALS leads
with accuracy from $0.955$ to $0.986$, compared with
$0.806$--$0.901$ for \tracemark. Recovery on ChordEdit remains harder for
\tracemark, with domain means of $0.230$--$0.427$.

\textbf{This advantage does not extend to baselines at their released
512-pixel resolution.} A four-feature control restores that carrier size for
VINE-R and Robust-Wide. Their mean accuracies over the three unseen channels rise from
$0.062$ to $0.997$ and from $0.165$ to $0.982$, respectively. TrustMark and
RoSteALS change little on those unseen channels when evaluated at 512 pixels.
\tracemark remains at 256 pixels, so this control does not measure its own
response to changing resolution, which would require retraining. Complete
channel-level results and edit-displacement measurements appear in
Appendix~\ref{app:resolution-domain}. RQ2 next isolates the benefit of training through
the editor within a shared architecture.

\subsection{Training Through the Editor (RQ2): Effect Across Attribution Methods}
\label{sec:exp:tracemark}
\label{sec:exp:training-readout}

\begin{figure}[t]
\centering
\includegraphics[width=\columnwidth]{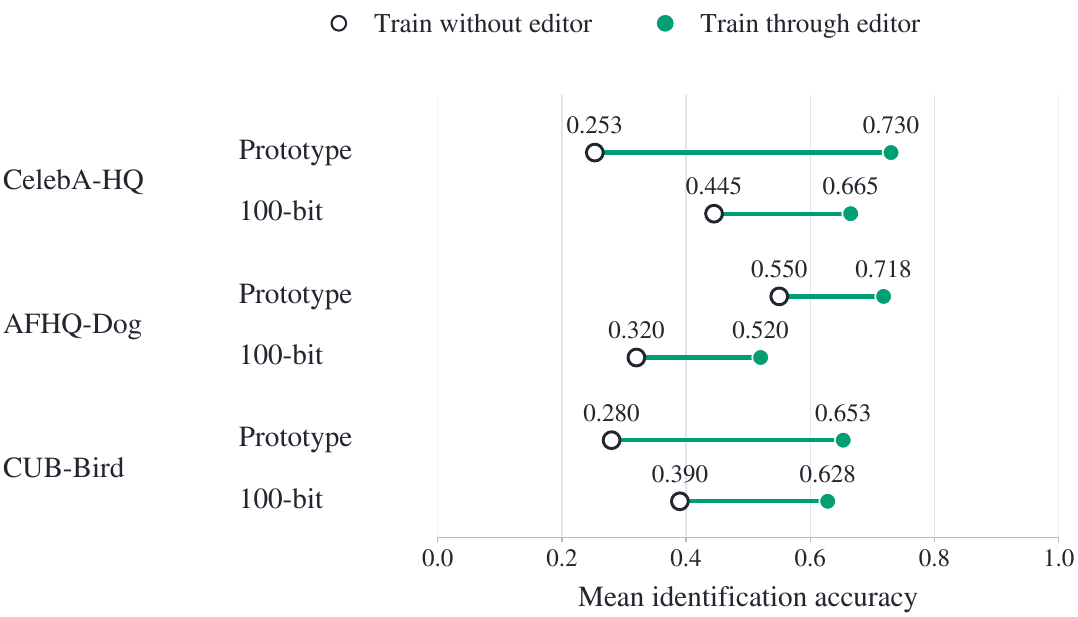}
\caption{Four-channel mean identification accuracy with and without
training through the editor, using two attribution methods in each domain.}
\label{fig:training-effect}
\end{figure}

\textbf{The system-level comparisons in RQ1 mix architecture and training
differences.} To isolate the effect of including the editor in training, we
construct a $2\times2$ control
that varies training through the editor and the attribution method on
blond hair (CelebA-HQ), angry expressions
(AFHQ-Dog), and crest (CUB-Bird). Within each feature, the four variants share
encoder and detector backbones, training data, number of training workers, random seed,
and $170$ training rounds, each containing the same image quota.
The editor-exposed variants train through
InstructPix2Pix, while the unexposed variants disable all edit-dependent
losses. Each condition uses either prototype matching or a $100$-bit decoder
with nearest-codeword attribution. Figure~\ref{fig:training-effect}
pairs the two training conditions for each attribution method and reports
mean accuracy over the four editing channels in
Table~\ref{tab:cross-editor-resolution}.

\begin{table}[t]
\centering
\caption{Post-edit identification accuracy after removing individual
objective terms in three image domains.}
\label{tab:loss-abl-main}
\footnotesize
\setlength{\tabcolsep}{3.8pt}
\begin{tabular*}{\columnwidth}{@{\extracolsep{\fill}}lcccc@{}}
\toprule
Domain & Full & $-\mathcal{L}_{\mathrm{car}}$ & $-\mathcal{L}_{\mathrm{cons}}$ & $-\mathcal{L}_{\mathrm{out}}$ \\
\midrule
CelebA-HQ & 0.927 & 0.944 & 0.963 & 0.743 \\
AFHQ-Dog  & 0.854 & 0.850 & 0.885 & 0.593 \\
CUB-Bird  & 0.818 & 0.839 & 0.905 & 0.613 \\
\midrule
Mean      & 0.866 & 0.878 & 0.918 & 0.650 \\
\bottomrule
\end{tabular*}
\end{table}

\textbf{Training through the editor improves all six tested combinations.}
Gains across features and attribution methods range from $0.167$ to $0.477$.
Mean gains are $0.339$ for
prototype matching and $0.219$ for binary decoding. The difference between
attribution methods, computed as prototype accuracy minus binary-decoding
accuracy, averages $0.096$ with editor exposure and $-0.024$ without it.
Positive values favor prototype matching, and the sign varies by feature.
In the AFHQ-Dog prototype variant, the $0.167$ training gain is smaller than
the prototype advantage over binary decoding, which is $0.198$ with editor
exposure and $0.230$ without it. Thus, training through the editor consistently benefits
both attribution methods in these controls, while the choice of attribution
method has a smaller average effect. Prototype matching simplifies registry
construction, but these results do not identify it as the source of edit
robustness.

\begin{figure*}[t]
\centering
\includegraphics[width=0.95\textwidth]{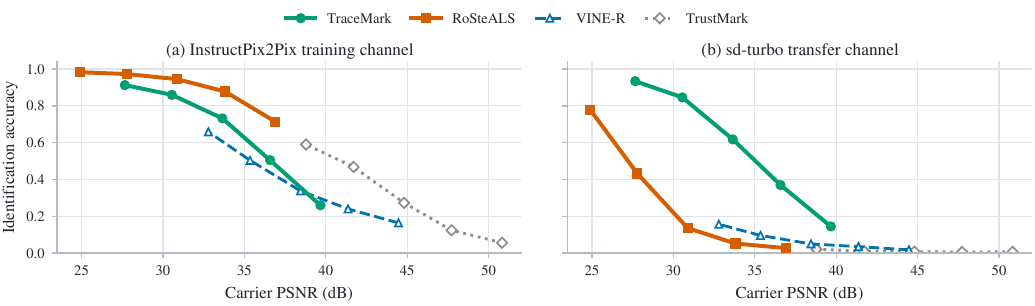}
\caption{Identification accuracy as a function of carrier peak signal-to-noise
ratio under InstructPix2Pix and sd-turbo editing.}
\label{fig:fidelity-frontier}
\end{figure*}

\begin{figure}[t]
\centering
\includegraphics[width=0.95\columnwidth]{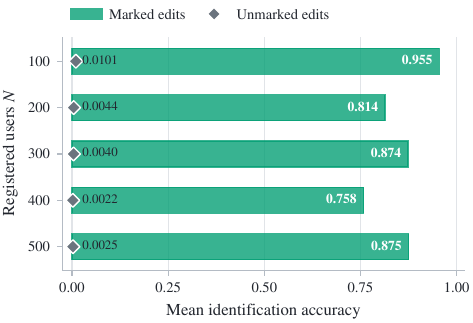}
\caption{Mean identification accuracy of population-specific models on marked
and unmarked edited images.}
\label{fig:population-capacity}
\end{figure}

\textbf{The training benefit emerges after editing.}
We also evaluate the same trained variants on their marked inputs before
editing and average each channel's accuracy over the
three features. Including the editor in training leaves pre-edit prototype accuracy
nearly unchanged ($0.983$ to $0.980$) and reduces binary accuracy ($0.983$ to
$0.877$). Through InstructPix2Pix, the corresponding accuracies rise from
$0.420$ to $0.920$ and from $0.410$ to $0.757$. All four variants can encode
user identity, but training through the editor improves its recovery
after transformation. On ChordEdit, the unexposed variants still average
$0.120$ and $0.123$, so some attribution persists without edit-dependent
training. The benefit is therefore associated with recovery after editing,
and the trained architecture also retains some capability without editor
exposure. Full per-channel
results and the $16$-bit width control appear in Appendix~\ref{app:training-readout}.

\textbf{Post-edit identification supervision provides the clearest accuracy benefit.}
The matched control changes all edit-dependent losses together. To examine
the contribution of individual terms, we next remove one loss at a time on one
feature from each domain. Table~\ref{tab:loss-abl-main} reports identification
accuracy. From these same runs, we also measure carrier $\ell_1$ distortion
and cosine similarity between embeddings of the marked input and its edited
output, with all three metrics reported in Appendix~\ref{app:objective-checks}.
Removing $\mathcal{L}_{\mathrm{out}}$ lowers accuracy by $0.184$--$0.261$ and
gives the lowest result in every domain. Removing
$\mathcal{L}_{\mathrm{cons}}$ lowers cosine alignment by $0.009$--$0.026$
while identification increases in all three domains. Removing
$\mathcal{L}_{\mathrm{car}}$ changes carrier $\ell_1$ by up to approximately
$0.001$ under the hard residual bound and produces mixed accuracy changes.
Identification also depends on competing-user scores, so better cosine
alignment alone need not improve accuracy, as the margin condition in
Section~\ref{sec:method:analysis} indicates.
Since $\mathcal{L}_{\mathrm{sim}}$ is retained, these ablations cannot isolate
the edit-fidelity term's contribution to post-edit identification accuracy.

\subsection{Operational Range (RQ3)}
\label{sec:exp:ablation}
\label{sec:exp:fidelity-control}

\textbf{The editor-dependent ordering persists at similar carrier fidelity.}
The methods have different carrier fidelity at their default strengths
(Appendix~\ref{app:baselines}). To compare accuracy at similar distortion levels,
we scale each marked-image residual with
$\beta\in\{0.35,0.5,0.7,1.0,1.4\}$ and plot carrier PSNR against post-edit
accuracy. Each point covers $90$ features and $60$ images per feature.
Near $33.7$ dB, RoSteALS reaches $0.878$ on InstructPix2Pix, compared with
$0.732$ for \tracemark (Fig.~\ref{fig:fidelity-frontier}(a)). On the unseen
sd-turbo channel, \tracemark instead reaches $0.618$, compared with $0.052$
for RoSteALS (Fig.~\ref{fig:fidelity-frontier}(b)). This ordering holds across
the plotted fidelity range. TrustMark and VINE-R also remain below
\tracemark on sd-turbo. Appendix~\ref{app:fidelity-sweep} provides the full sweep.

\textbf{Similar edit magnitudes can produce different attribution outcomes.}
Section~\ref{sec:motivation:backbone} compares IP2P and SDXL-Turbo at nearly
equal latent displacement. Under that $90$-feature protocol, \tracemark
reaches $0.866$ on IP2P and
$0.430$ on SDXL-Turbo. On the latter it exceeds RoSteALS at $0.249$, but falls
below VINE-R at $0.922$ and Robust-Wide at $0.648$. RQ1 and Appendix~\ref{app:resolution-domain} report the complementary carrier-resolution control for the
released baseline models.

\begin{table}[t]
\centering
\caption{Cross-instruction identification with fixed watermark models.
Each source model is tested on three target instructions.}
\label{tab:cross-instr-main}
\footnotesize
\setlength{\tabcolsep}{3pt}
\begin{tabular*}{\columnwidth}{@{\extracolsep{\fill}}lccc@{}}
\toprule
Training instruction & Marked mean & Marked range & Unmarked mean \\
\midrule
\multicolumn{4}{l}{\textit{CelebA-HQ}} \\
Add glasses & 0.843 & 0.680--0.975 & 0.003 \\
Make smiling & 0.613 & 0.120--0.905 & 0.003 \\
Make anime & 0.858 & 0.735--0.920 & 0.023 \\
\midrule
\multicolumn{4}{l}{\textit{AFHQ-Dog}} \\
Turn into cat & 0.540 & 0.365--0.805 & 0.015 \\
Turn into lion & 0.425 & 0.250--0.725 & 0.003 \\
Make anime & 0.305 & 0.190--0.485 & 0.012 \\
\midrule
\multicolumn{4}{l}{\textit{CUB-Bird}} \\
Make head red & 0.740 & 0.630--0.870 & 0.003 \\
Make tail blue & 0.757 & 0.665--0.855 & 0.010 \\
\midrule
All 24 instruction pairs & 0.635 & 0.120--0.975 & 0.009 \\
\bottomrule
\end{tabular*}
\end{table}

\textbf{User attribution can survive a change of instruction without retraining.}
The previous tests retain each model's training instruction. To examine
whether a task-specific model can also serve other requests, we keep
InstructPix2Pix fixed and change only the instruction supplied at evaluation.
We use eight trained models across the three domains and test each on three
other instructions from the same domain, with $200$ covers per pair. The
source model embeds the user signal before the new edit, and its unchanged
detector and prototypes identify the user afterward. Paired unmarked inputs
receive the same edit and editor seed. Table~\ref{tab:cross-instr-main}
summarizes the $24$ instruction pairs, with mean accuracy of $0.635$ on marked
outputs and $0.009$ on unmarked outputs. Recovery varies substantially with
the target instruction. For example, the glasses-trained model reaches
$0.975$ for smiling but $0.680$ for anime. The smiling-trained model reaches
only $0.120$ for a Van Gogh-style edit. These results demonstrate reuse for
some requests beyond the training instruction, but do not establish reliable
recovery for arbitrary instructions. Appendix~\ref{app:cross-instr} lists every
source-target pair and its marked and unmarked accuracy.

\label{sec:exp:population-capacity}

\textbf{Separately trained models support registries of up to $500$ users.}
Figure~\ref{fig:population-capacity} reports results for $100$--$500$ users
across four CelebA-HQ tasks. Mean accuracy ranges from $0.758$ to $0.955$,
with $0.875$ at $N=500$.
Applying each detector to edited images without an embedded watermark gives
mean accuracies from $0.0022$ to $0.0101$, close to the chance level $1/N$.
This unmarked control covers all four tasks except at $N=200$, where paired
records are available for two tasks.
The near-chance unmarked results support the contribution of the embedded
user signal to identification.
Optimization budgets and per-task measurements appear in Appendix~\ref{app:population-capacity} for each of the five evaluated registry sizes.

\section{Conclusion}
\label{sec:conclusion}

We presented \tracemark for tracing edited images to the users who requested
them. Joint training of user-conditioned watermark embedding and detection
through a fixed editor improves attribution in matched controls across
three domains, and separately optimized models support registries of up to
$500$ users. For an editing service, the resulting image-based signal can
reconnect redistributed content with its request records when identifying
metadata is unavailable. This link supports provenance tracking beyond the
service interface under trusted user assignment. The findings also make
exact user recovery a concrete target for watermark training and evaluation:
for user attribution, a recovered signal must distinguish the intended user
from others in the registry. Editor-aware training helps achieve this
within the tested settings, while comparisons with released watermarks
identify conditions in which existing methods already perform well.
These comparisons guide the choice between released watermarks and
task-specific training, with validation needed for the intended editor,
carrier resolution, and image population.

\bibliographystyle{IEEEtran}
\bibliography{refs}

\clearpage
\twocolumn[\section*{Appendices}]
\appendices

\section{Datasets and Editing Features}
\label{app:data}
We evaluate faces from high-quality CelebA
(CelebA-HQ)~\cite{karras2018progressive}, animal faces from the Animal
Faces-HQ (AFHQ) Dog split~\cite{choi2020stargan}, and fine-grained birds
from Caltech-UCSD Birds-200-2011 (CUB-200-2011)~\cite{wah2011cub}. Their
training pools contain $28{,}000$,
$4{,}739$, and $5{,}994$ images. CelebA-HQ and AFHQ-Dog are center-cropped.
CUB bounding boxes are padded by $10\%$ before cropping. All images are
resized to $256\times256$. Each domain has $30$ short editing instructions.
They cover identities, expressions, attributes, and styles for faces, breed,
color, style, and action for dogs, and anatomical attributes for birds.

\section{Baseline Configurations}
\label{app:baselines}
We compare \tracemark with four released checkpoints. TrustMark, RoSteALS,
and VINE-R receive one $100$-bit codeword per user from the balanced greedy
codebook used in the main paper. Robust-Wide retains its native $64$-bit
message. Attribution selects the registered codeword with the fewest bit
differences from the decoded message. Equal-distance matches count as
identification errors.

\textbf{TrustMark}~\cite{bui2023trustmark} uses its released \texttt{Q} model
with a $100$-bit raw message, error-correcting code (ECC) disabled, and
watermark strength $1.0$.
\textbf{RoSteALS}~\cite{bui2023rosteals} uses the public \texttt{vq-f4}
first-stage backbone and released
\texttt{epoch=000017-step=000449999.ckpt} checkpoint.
\textbf{VINE-R}~\cite{lu2024vine} uses the released
\texttt{Shilin-LU/VINE-R-Enc} encoder and
\texttt{Shilin-LU/VINE-R-Dec} decoder.
\textbf{Robust-Wide}~\cite{hu2024robustwide} uses its released
$64$-bit encoder and decoder. Both VINE-R and Robust-Wide produce native
$512\times512$ carriers. The common-resolution experiment resizes these to
$256\times256$, while the resolution control retains the native size through
editing and decoding.

Every baseline uses the same covers, instructions, and editor seeds. Its
decoder is evaluated on the resulting edited images.
Carrier fidelity uses peak signal-to-noise ratio (PSNR), structural similarity
index measure (SSIM)~\cite{wang2004ssim}, and learned perceptual image patch
similarity (LPIPS)~\cite{zhang2018lpips}.

\begin{table}[t]
\centering
\caption{Carrier fidelity at each method's native watermark strength over
$90$ semantic editing features.}
\label{tab:carrier-quality-supp}
\setlength{\tabcolsep}{5pt}
\begin{tabular}{lccc}
\toprule
Method & PSNR $\uparrow$ & SSIM $\uparrow$ & LPIPS $\downarrow$ \\
\midrule
\tracemark & 30.39 & 0.9566 & 0.0783 \\
TrustMark & 42.08 & 0.9883 & 0.0010 \\
RoSteALS & 29.07 & 0.8599 & 0.0348 \\
VINE-R & 36.53 & 0.9833 & 0.0089 \\
\bottomrule
\end{tabular}
\end{table}

\section{Carrier-Strength Sweep}
\label{app:fidelity-sweep}

Table~\ref{tab:fidelity-sweep-supp} gives the complete measurements behind
the main-paper carrier-strength figure. Each entry averages $60$ images for
each of $90$ semantic editing features. We multiply the difference between
the marked and original image by $\beta$ before editing, with $\beta=1$
retaining the original watermark strength. The carrier PSNR is measured
before editing.

\begin{table}[t]
\centering
\caption{Carrier PSNR and post-edit identification accuracy across watermark
strengths.}
\label{tab:fidelity-sweep-supp}
\footnotesize
\setlength{\tabcolsep}{3pt}
\begin{tabular*}{\columnwidth}{@{\extracolsep{\fill}}lccccc@{}}
\toprule
$\beta$ & $0.35$ & $0.50$ & $0.70$ & $1.00$ & $1.40$ \\
\midrule
\multicolumn{6}{l}{\emph{Carrier PSNR (dB)}} \\
\tracemark & 39.67 & 36.58 & 33.65 & 30.55 & 27.66 \\
TrustMark & 50.83 & 47.73 & 44.81 & 41.71 & 38.79 \\
RoSteALS & 36.90 & 33.81 & 30.88 & 27.79 & 24.89 \\
VINE-R & 44.47 & 41.37 & 38.45 & 35.35 & 32.79 \\
\midrule
\multicolumn{6}{l}{\emph{Identification accuracy: InstructPix2Pix}} \\
\tracemark & 0.260 & 0.505 & 0.732 & 0.860 & 0.913 \\
TrustMark & 0.056 & 0.124 & 0.273 & 0.468 & 0.590 \\
RoSteALS & 0.714 & 0.878 & 0.945 & 0.973 & 0.983 \\
VINE-R & 0.164 & 0.240 & 0.338 & 0.504 & 0.659 \\
\midrule
\multicolumn{6}{l}{\emph{Identification accuracy: sd-turbo, $s=0.2$}} \\
\tracemark & 0.145 & 0.370 & 0.618 & 0.846 & 0.934 \\
TrustMark & 0.007 & 0.006 & 0.008 & 0.011 & 0.021 \\
RoSteALS & 0.027 & 0.052 & 0.136 & 0.432 & 0.779 \\
VINE-R & 0.018 & 0.035 & 0.050 & 0.096 & 0.157 \\
\bottomrule
\end{tabular*}
\end{table}

\section{Carrier-Resolution and Domain Breakdowns}
\label{app:resolution-domain}

The main paper summarizes the carrier-resolution control. This section gives
the complete channel-level identification results, the corresponding edit
displacement measurements, and the per-domain \tracemark breakdown.

\begin{table}[t]
\centering
\caption{Four-feature identification accuracy at 256- and 512-pixel carrier
resolutions. IP2P denotes InstructPix2Pix.}
\label{tab:resolution-identification-full}
\footnotesize
\setlength{\tabcolsep}{2pt}
\begin{tabular*}{\columnwidth}{@{\extracolsep{\fill}}lcccccc@{}}
\toprule
Method & Pixels & Identity & IP2P &
\shortstack{sd-turbo\\$s=0.2$} &
\shortstack{sd-turbo\\p2p-style\\$s=0.3$} & ChordEdit \\
\midrule
\tracemark & 256 & 0.985 & 0.907 & 0.873 & 0.780 & 0.287 \\
\midrule
\multirow{2}{*}{TrustMark}
 & 256 & 1.000 & 0.557 & 0.020 & 0.007 & 0.015 \\
 & 512 & 1.000 & 0.988 & 0.035 & 0.013 & 0.003 \\
\midrule
\multirow{2}{*}{RoSteALS}
 & 256 & 1.000 & 0.995 & 0.420 & 0.138 & 0.083 \\
 & 512 & 1.000 & 0.990 & 0.425 & 0.133 & 0.083 \\
\midrule
\multirow{2}{*}{VINE-R}
 & 256 & 1.000 & 0.605 & 0.100 & 0.062 & 0.025 \\
 & 512 & 1.000 & 1.000 & 1.000 & 0.998 & 0.993 \\
\midrule
\multirow{2}{*}{Robust-Wide}
 & 256 & 1.000 & 0.975 & 0.297 & 0.147 & 0.050 \\
 & 512 & 1.000 & 1.000 & 1.000 & 0.993 & 0.953 \\
\bottomrule
\end{tabular*}
\end{table}

The resolution control also measures each editor on unmarked covers. The same
four features, covers, instructions, and seeds are used at both resolutions.
Normalized latent displacement is measured with the fixed Stable Diffusion
1.5 variational autoencoder used for the main-paper edit-magnitude control.

\begin{table}[t]
\centering
\caption{Edit displacement at carrier resolutions of 256 and 512 pixels.}
\label{tab:resolution-displacement}
\setlength{\tabcolsep}{5pt}
\begin{tabular}{lccc}
\toprule
Editing channel & 256 & 512 & Ratio \\
\midrule
InstructPix2Pix & 0.513 & 0.436 & 0.85 \\
sd-turbo, $s=0.2$ & 0.474 & 0.370 & 0.78 \\
sd-turbo, p2p-style $s=0.3$ & 0.506 & 0.400 & 0.79 \\
ChordEdit & 0.869 & 0.557 & 0.64 \\
\bottomrule
\end{tabular}
\end{table}

Table~\ref{tab:resolution-displacement} shows that the 512-pixel pipelines
change the covers less, with ratios of 512-pixel to 256-pixel displacement
from $0.64$ to $0.85$. Resolution thus changes both the watermark input and
the editing operation. On the three unseen editing channels, VINE-R and
Robust-Wide improve to at least $0.953$, while TrustMark and RoSteALS remain
nearly unchanged. TrustMark does improve substantially on InstructPix2Pix.
These different responses show that the effect of resolution is
method-dependent. No 512-pixel \tracemark model is evaluated here.

\begin{figure}[t]
\centering
\includegraphics[width=0.9\columnwidth]{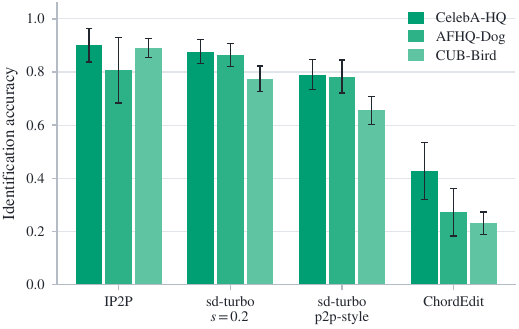}
\caption{\tracemark identification accuracy by image domain and editing
channel.}
\label{fig:domain-transfer}
\end{figure}

Figure~\ref{fig:domain-transfer} separates the 90-feature \tracemark results
by dataset. Both sd-turbo configurations retain higher accuracy than
ChordEdit in every domain, so the aggregate ranking also holds within each
dataset.

\section{Implementation and Reproducibility}
\label{app:impl}
Table~\ref{tab:hparams} lists the main hyperparameters, with architecture,
training, and evaluation details below. The implementation
is PyTorch~$2.6$ with \texttt{diffusers}~$0.27$, and one model is
trained per editing feature with no weight sharing across features.
Within each model, all registered users share the encoder and detector
weights, with separate user embeddings and prototypes.

\paragraph{Identifier encoder}
The encoder $E_\theta$ broadcasts a learned $d_u{=}64$ user embedding over
the spatial grid and concatenates it with the three image channels, forming
a $67$-channel input. Four convolutional blocks
(convolution, GroupNorm, SiLU) of width $96$ process it, and a final
$1\times 1$ convolution produces a $3$-channel residual $\delta$.
The residual $\alpha\tanh\delta$, with fixed $\alpha{=}0.07$, is added to the
cover before clipping to $[-1,1]$.

\paragraph{Detector and prototype bank}
The detector $D_\phi$ is a five-stage strided convolutional backbone
with channel widths $64\!\to\!96\!\to\!128\!\to\!192\!\to\!256$,
followed by global average pooling and a linear projection to a
$d{=}128$ dimensional embedding. This embedding and the learned per-user
prototypes are $\ell_2$-normalized. Class logits are their cosine similarities
multiplied by a learned positive scale, initialized to $16$ and capped at
$100$. The detector is shared before and after editing.

\paragraph{Optimization and edit curriculum}
The encoder, detector, and prototypes are trained jointly with Adam at
learning rate $1.8\times 10^{-4}$, cosine-annealed to $5\%$ of its
initial value over $T_{\mathrm{train}}{=}170$ training rounds, using
DistributedDataParallel across $6$ A100-80GB graphics processing units
(GPUs) with a global batch of $24$ images ($4$ per GPU). Each round
processes a fixed quota of
$1{,}392$ images per feature with random horizontal flips and assigns
user indices by the deterministic mapping
$u=(t\,N_{\mathrm{img}}+i)\bmod N$. Here the implementation uses zero-based
user, round, and sample indices, and $N_{\mathrm{img}}=1{,}392$ is the
per-round image quota. This assignment updates the $N{=}100$ prototypes
near-uniformly. The first $t_{\mathrm{warm}}{=}35$ rounds use marked but
unedited images ($r_t{=}0$). The editor is then included, and the weights of
the three edit-dependent losses increase linearly to their full values over
$t_{\mathrm{ramp}}{=}90$ rounds, remaining fixed thereafter. Wall-clock time
is about $4.5$\,h per feature. The full study of $90$ features takes about
$405$ server hours, equal to roughly $2{,}430$ A100 GPU-hours with six
devices per run.

\paragraph{Editor invocation}
The frozen editor is run with the
\texttt{EulerAncestralDiscreteScheduler} for $6$ inference steps at
text guidance scale $7.0$ and image guidance scale $2.0$. Each
watermarked cover and its clean counterpart use the same per-image seed,
so the edit-fidelity comparison holds sampling randomness fixed. Gradients
reach the encoder through the editor without updating the editor weights.

\paragraph{Evaluation protocol}
The trained-editor and cross-editor comparisons use $100$ images per feature
across all $90$ features. The equal-displacement study uses all $90$ features
and $50$ images per feature. The carrier-resolution control uses four features
and $100$ images per feature. The carrier-strength sweep uses all $90$ features and
$60$ images per feature. The population study uses four CelebA-HQ tasks at
$N\in\{100,200,300,400,500\}$ with the training-round and worker budgets reported
in Table~\ref{tab:population-capacity-supp}. Seeds are fixed within each paired
comparison. The \tracemark carriers used in these evaluations are selected
from outputs saved during the final training round on each training pool.
The baselines receive the same covers in paired comparisons. The current
protocol does not evaluate a disjoint cover split.
Identification accuracy uses nearest-prototype matching for
\tracemark and nearest-codeword matching for the bitstring methods.

\paragraph{Detection scoring and model size}
For watermark-presence detection, we evaluated four confidence scores: maximum
softmax probability, the top-$1$ versus top-$2$ logit margin, the raw maximum
logit $\max_v\boldsymbol{\ell}_v$, and negative softmax entropy. Averaged over
the $90$ editing features, their mean areas under the receiver operating
characteristic curve (AUCs) are $0.884$, $0.834$, $0.946$, and $0.909$ in the
same order. These scores assess watermark presence separately from user
identification. At $N{=}100$
users the encoder $E_\theta$ has $3.1\times 10^5$ and the detector
$D_\phi$ has $8.8\times 10^5$ parameters.

\begin{table}[t]
\centering
\caption{Full \tracemark configuration for the paper-scale runs.}
\label{tab:hparams}
\setlength{\tabcolsep}{5pt}
\begin{tabular}{ll}
\toprule
Setting & Value \\
\midrule
User embedding dim $d_u$              & $64$ \\
Encoder blocks / width               & $4$ / $96$ \\
Carrier strength $\alpha$            & $0.07$ \\
Detector backbone widths             & $64,96,128,192,256$ \\
Embedding dim $d$                    & $128$ \\
Logit scale (init / cap)             & $16$ / $100$ \\
\midrule
Optimiser                            & Adam \\
Learning rate                        & $1.8\times10^{-4}$ \\
Learning-rate schedule               & cosine to $5\%$ \\
Training rounds $T_{\mathrm{train}}$ & $170$ \\
Warm-up / ramp (rounds) & $35$ / $90$ \\
GPUs / global batch                  & $6\times$A100 / $24$ \\
Images per feature / users $N$       & $1{,}392$ / $100$ \\
\midrule
Editor steps                         & $6$ \\
Text / image guidance               & $7.0$ / $2.0$ \\
Resolution                           & $256\times256$ \\
Loss weights $\lambda_{\mathrm{id/sim/car/cons}}$ & $3.2, 2.0, 0.85, 0.8$ \\
Edited-image weight $\lambda_{\mathrm{out}}$      & $1.4$ \\
Seed                                 & $1234$ \\
\bottomrule
\end{tabular}
\end{table}

\section{Training Exposure and Attribution Methods}
\label{app:training-readout}

Table~\ref{tab:training-readout-full} gives the exact results summarized by
the main-paper training-control figure. Within each feature, the four variants
share the training data, random seed, GPU worker count, training-round budget,
and encoder and detector backbones. The upper block reports the four-channel
mean by feature. The lower block reports each channel after averaging over
the three features. Exposed variants include InstructPix2Pix in training,
while unexposed variants disable all edit-dependent losses. The Training
columns subtract unexposed accuracy from exposed accuracy. The Method effect
columns subtract binary-decoding accuracy from prototype-matching accuracy.
The identity channel evaluates the same models on marked inputs before editing.

\begin{table*}[t]
\centering
\caption{Identification accuracy for the training-exposure and
attribution-method control by image domain and editing channel.}
\label{tab:training-readout-full}
\footnotesize
\setlength{\tabcolsep}{3.7pt}
\begin{tabular*}{\textwidth}{@{\extracolsep{\fill}}lcccccccc@{}}
\toprule
& \multicolumn{3}{c}{Prototype matching} &
  \multicolumn{3}{c}{100-bit decoding} &
  \multicolumn{2}{c}{Method effect} \\
\cmidrule(lr){2-4}\cmidrule(lr){5-7}\cmidrule(l){8-9}
Feature & Exposed & Unexposed & Training & Exposed & Unexposed & Training & Exposed & Unexposed \\
\midrule
Blond hair, CelebA-HQ & 0.730 & 0.253 & $+0.477$ & 0.665 & 0.445 & $+0.220$ & $+0.065$ & $-0.192$ \\
Angry, AFHQ-Dog & 0.718 & 0.550 & $+0.167$ & 0.520 & 0.320 & $+0.200$ & $+0.198$ & $+0.230$ \\
Crest, CUB-Bird & 0.653 & 0.280 & $+0.373$ & 0.628 & 0.390 & $+0.238$ & $+0.025$ & $-0.110$ \\
\midrule
Mean & 0.700 & 0.361 & $+0.339$ & 0.604 & 0.385 & $+0.219$ & $+0.096$ & $-0.024$ \\
\bottomrule
\end{tabular*}
\vspace{5pt}
\begin{tabular*}{\textwidth}{@{\extracolsep{\fill}}lcccc@{}}
\toprule
Channel & Exposed, prototype & Exposed, 100-bit & Unexposed, prototype & Unexposed, 100-bit \\
\midrule
Identity & 0.980 & 0.877 & 0.983 & 0.983 \\
Variational autoencoder round trip & 0.950 & 0.830 & 0.623 & 0.717 \\
InstructPix2Pix & 0.920 & 0.757 & 0.420 & 0.410 \\
sd-turbo, $s=0.2$ & 0.857 & 0.763 & 0.473 & 0.560 \\
sd-turbo, p2p-style $s=0.3$ & 0.770 & 0.673 & 0.430 & 0.447 \\
ChordEdit & 0.253 & 0.223 & 0.120 & 0.123 \\
\bottomrule
\end{tabular*}
\end{table*}

All four variants retain mean identity-channel accuracy between $0.877$ and
$0.983$. Training through the editor improves recovery after editing for
both attribution methods in every domain.

\begin{figure}[t]
\centering
\includegraphics[width=0.9\columnwidth]{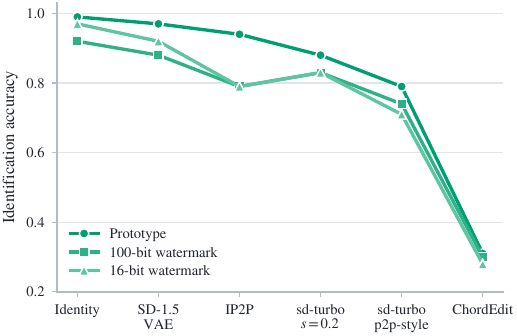}
\caption{Identification accuracy of editor-exposed variants with prototype
matching, 16-bit decoding, and 100-bit decoding across six channels.}
\label{fig:readout-control-supp}
\end{figure}

Figure~\ref{fig:readout-control-supp} supplies the additional watermark-width
control. The $16$-bit and $100$-bit variants both reach $0.790$ on
InstructPix2Pix. Their ChordEdit accuracies are $0.280$ and $0.300$. Watermark
width changes accuracy on unedited marked inputs, while the two binary
decoders have similar accuracies on the tested editing channels.

\section{Cross-Instruction Reuse}
\label{app:cross-instr}
The main paper summarizes reuse of eight task-specific models across $24$
source-target instruction pairs. Three source models use CelebA-HQ, three
use AFHQ-Dog, and two use CUB-Bird. For each model, three target instructions
from the same domain are evaluated without retraining. Each pair uses $200$
covers sampled from that source model's retained image pool, with the user
identifiers stored by its checkpoint. The source encoder marks each cover,
InstructPix2Pix applies the target instruction, and the source detector and
prototype bank predict the user. For the unmarked control, the same cover
receives the same instruction and editor seed without watermark embedding.
Identification accuracy is computed against the same assigned user in both
conditions. These tests change the instruction and do not measure
generalization to held-out images.

\begin{table}[t]
\centering
\caption{Identification accuracy for all cross-instruction pairs, with
200 images per pair.}
\label{tab:cross-instr}
\footnotesize
\setlength{\tabcolsep}{3pt}
\begin{tabular*}{\columnwidth}{@{\extracolsep{\fill}}llcc@{}}
\toprule
Training instruction & Test instruction & Marked & Unmarked \\
\midrule
\multicolumn{4}{l}{\textit{CelebA-HQ}} \\
\multirow{3}{*}{Add glasses}
 & Make older & 0.875 & 0.005 \\
 & Make smiling & 0.975 & 0.000 \\
 & Make anime & 0.680 & 0.005 \\
\multirow{3}{*}{Make smiling}
 & Add glasses & 0.905 & 0.000 \\
 & Make older & 0.815 & 0.005 \\
 & Van Gogh style & 0.120 & 0.005 \\
\multirow{3}{*}{Make anime}
 & Add glasses & 0.920 & 0.020 \\
 & Add makeup & 0.920 & 0.025 \\
 & Add beard & 0.735 & 0.025 \\
\midrule
\multicolumn{4}{l}{\textit{AFHQ-Dog}} \\
\multirow{3}{*}{Turn into cat}
 & Turn into lion & 0.365 & 0.020 \\
 & Make anime & 0.805 & 0.010 \\
 & Make red & 0.450 & 0.015 \\
\multirow{3}{*}{Turn into lion}
 & Make anime & 0.725 & 0.010 \\
 & Make blue & 0.300 & 0.000 \\
 & Make purple & 0.250 & 0.000 \\
\multirow{3}{*}{Make anime}
 & Turn into cat & 0.485 & 0.030 \\
 & Turn into lion & 0.190 & 0.005 \\
 & Turn into wolf & 0.240 & 0.000 \\
\midrule
\multicolumn{4}{l}{\textit{CUB-Bird}} \\
\multirow{3}{*}{Make head red}
 & Make belly yellow & 0.720 & 0.000 \\
 & Make crown black & 0.630 & 0.005 \\
 & Make breast red & 0.870 & 0.005 \\
\multirow{3}{*}{Make tail blue}
 & Make breast orange & 0.665 & 0.010 \\
 & Make beak longer & 0.855 & 0.005 \\
 & Make head red & 0.750 & 0.015 \\
\bottomrule
\end{tabular*}
\end{table}

Table~\ref{tab:cross-instr} reports all pairs, weighted equally when computing
means. Domain means are $0.772$ for CelebA-HQ, $0.423$ for AFHQ-Dog, and
$0.748$ for CUB-Bird. These tests measure instruction reuse without isolating
the cause of individual accuracy drops.

\section{Population-Capacity Measurements}
\label{app:population-capacity}

Table~\ref{tab:population-capacity-supp} records the values and optimization
budgets summarized by the main-paper population figure. The $N=200$ glasses
and makeup runs use eight workers. The red-hair and smiling runs use three.
The $N=300$ row uses six to eight workers across tasks. All other rows use six.
The unmarked-edit column averages the same four tasks except at $N=200$, where
only the glasses and makeup paired records were retained.

\begin{table}[t]
\centering
\caption{Identification accuracy and optimization budgets for four
CelebA-HQ tasks with registries from 100 to 500 users.}
\label{tab:population-capacity-supp}
\footnotesize
\setlength{\tabcolsep}{3pt}
\begin{tabular*}{\columnwidth}{@{\extracolsep{\fill}}lccccc@{}}
\toprule
Registry size $N$ & 100 & 200 & 300 & 400 & 500 \\
\midrule
Training rounds & 170 & 200 & 300 & 400 & 700 \\
Workers & 6 & 8 / 3 & 6--8 & 6 & 6 \\
\midrule
Glasses & 0.967 & 0.891 & 0.916 & 0.782 & 0.870 \\
Makeup & 0.952 & 0.876 & 0.891 & 0.700 & 0.869 \\
Red hair & 0.939 & 0.754 & 0.851 & 0.718 & 0.900 \\
Smiling & 0.961 & 0.735 & 0.837 & 0.831 & 0.861 \\
Mean & 0.955 & 0.814 & 0.874 & 0.758 & 0.875 \\
Unmarked & 0.0101 & 0.0044 & 0.0040 & 0.0022 & 0.0025 \\
\bottomrule
\end{tabular*}
\end{table}

\section{Objective Ablations}
\label{app:objective-checks}

We remove the carrier penalty
$\mathcal{L}_{\mathrm{car}}$, embedding-consistency penalty
$\mathcal{L}_{\mathrm{cons}}$, or post-edit identification term
$\mathcal{L}_{\mathrm{out}}$ one at a time on one feature per domain.
Table~\ref{tab:loss-abl} reports
identification accuracy, carrier $\ell_1$ distortion, and cosine similarity
between pre-edit and post-edit embeddings. The hard residual bound remains
active in every variant. Removing $\mathcal{L}_{\mathrm{out}}$ gives the
lowest accuracy in each domain. Removing $\mathcal{L}_{\mathrm{cons}}$ lowers
cosine alignment despite higher identification accuracy.

\begin{table}[t]
\centering
\caption{Loss ablation in three image domains.}
\label{tab:loss-abl}
\footnotesize
\setlength{\tabcolsep}{3.8pt}
\begin{tabular*}{\columnwidth}{@{\extracolsep{\fill}}llcccc@{}}
\toprule
Metric & Domain & Full & no $\mathcal{L}_{\mathrm{car}}$ & no $\mathcal{L}_{\mathrm{cons}}$ & no $\mathcal{L}_{\mathrm{out}}$ \\
\midrule
\multirow{3}{*}{Accuracy $\uparrow$}
 & CelebA-HQ & 0.927 & 0.944 & 0.963 & \textbf{0.743} \\
 & AFHQ-Dog  & 0.854 & 0.850 & 0.885 & \textbf{0.593} \\
 & CUB-Bird  & 0.818 & 0.839 & 0.905 & \textbf{0.613} \\
\midrule
\multirow{3}{*}{Carrier $\ell_1$ $\downarrow$}
 & CelebA-HQ & 0.030 & 0.031 & 0.031 & 0.030 \\
 & AFHQ-Dog  & 0.031 & 0.031 & 0.031 & 0.031 \\
 & CUB-Bird  & 0.028 & 0.029 & 0.026 & 0.027 \\
\midrule
\multirow{3}{*}{Cosine $\uparrow$}
 & CelebA-HQ & 0.952 & 0.932 & 0.943 & \textbf{0.902} \\
 & AFHQ-Dog  & 0.933 & 0.922 & 0.912 & \textbf{0.873} \\
 & CUB-Bird  & 0.874 & 0.859 & 0.848 & \textbf{0.791} \\
\bottomrule
\end{tabular*}
\end{table}

\end{document}